\documentclass[10pt,twocolumn,letterpaper]{article}

\usepackage[pagenumbers]{cvpr} 

\usepackage{graphicx}
\usepackage{amsmath}
\usepackage{amssymb}
\usepackage{booktabs}

\usepackage{times}
\usepackage{epsfig}
\usepackage{graphicx}
\usepackage{amsmath}
\usepackage{amssymb}
\usepackage{algorithm}
\usepackage{algpseudocode}
\usepackage{threeparttable}
\usepackage{multirow}
\usepackage{booktabs}
\usepackage{caption}
\usepackage{mathrsfs}
\usepackage{bbding}
\usepackage[accsupp]{axessibility}

\usepackage[pagebackref,breaklinks,colorlinks]{hyperref}

\usepackage[capitalize]{cleveref}
\crefname{section}{Sec.}{Secs.}
\Crefname{section}{Section}{Sections}
\Crefname{table}{Table}{Tables}
\crefname{table}{Tab.}{Tabs.}

\def\cvprPaperID{19} 
\def\confName{CVPR}
\def\confYear{2022}

\begin{document}
	
	\title{Compositional Temporal Grounding\\ with Structured Variational Cross-Graph Correspondence Learning}
	

\author{Juncheng Li$~\textsuperscript{\rm 1}$ \and Junlin Xie$~\textsuperscript{\rm 1}$  \and  Long Qian$~\textsuperscript{\rm 1}$ \and  Linchao Zhu$~\textsuperscript{\rm 2}$   \and  Siliang Tang$~\textsuperscript{\rm 1}$  \and   Fei Wu$~\textsuperscript{\rm 1}$ \and Yi Yang$~\textsuperscript{\rm 1}$  \and Yueting Zhuang$~\textsuperscript{\rm 1}$\thanks{Yueting Zhuang is the corresponding author.}  \and Xin Eric Wang$~\textsuperscript{\rm 3}$ \and \vspace{-0.4cm}\\
	\small{$~\textsuperscript{\rm 1}$ Zhejiang University}, 
	\small{$~\textsuperscript{\rm 2}$ University of Technology Sydney},
	\small{$~\textsuperscript{\rm 3}$ University of California, Santa Cruz}
	\\{\tt\small {\{junchengli, 22051289, qianlong0926, siliang, wufei, yangyics, yzhuang\}}@zju.edu.cn}
	\\{\tt\small linchao.zhu@uts.edu.au, xwang366@ucsc.edu}
}

	\maketitle
	
	\begin{abstract}
     	Temporal grounding in videos aims to localize one target video segment that semantically corresponds to a given query sentence.  Thanks to the semantic diversity of natural language descriptions, temporal grounding allows activity grounding beyond pre-defined classes and has received increasing attention in recent years. The semantic diversity is rooted in the principle of compositionality in linguistics, where novel semantics can be systematically described by combining known words in novel ways~(\textbf{compositional generalization}). However, current temporal grounding datasets do not specifically test for the compositional generalizability. To systematically measure the compositional generalizability of temporal grounding models, we introduce a new Compositional Temporal Grounding task and construct two new dataset splits, \ie, Charades-CG and ActivityNet-CG. Evaluating the state-of-the-art methods on our new dataset splits, we empirically find that they fail to generalize to queries with novel combinations of seen words. To tackle this challenge, we propose a  variational cross-graph reasoning framework that explicitly decomposes video and language into multiple structured hierarchies and learns fine-grained semantic correspondence among them. Experiments illustrate the superior compositional generalizability of our approach. The repository of this work is at \url{https://github.com/YYJMJC/Compositional-Temporal-Grounding}.

	\end{abstract}
	
	
	\vspace{-0.5cm}
	\section{Introduction}
	Understanding rich and diverse activities in videos is a prominent and fundamental goal of video understanding. While there have been significant works in activity recognition~\cite{carreira2017quo, feichtenhofer2016convolutional} and localization~\cite{ma2016learning, singh2016multi}, one major limitation of these works is that they are restricted to pre-defined action classes, thus suffering from scaling to various complex activities. A natural solution to this problem is to utilize the systematic compositionality~\cite{fodor1988connectionism, chomsky2009syntactic, montague1970universal} of human language, which allows us to form novel compositions by combining known words in novel ways to describe unseen activities~(\ie \textbf{compositional generalization}). Therefore, a new task, namely temporal grounding in videos~\cite{gao2017tall, krishna2017dense}, has recently received increasing attention. Formally, give an untrimmed video and a query sentence, it aims to identify the start and end timestamps of one specific moment that semantically corresponds to the given query sentence.
	
	\begin{figure}[!t]
		\centering
		\includegraphics[width=\linewidth]{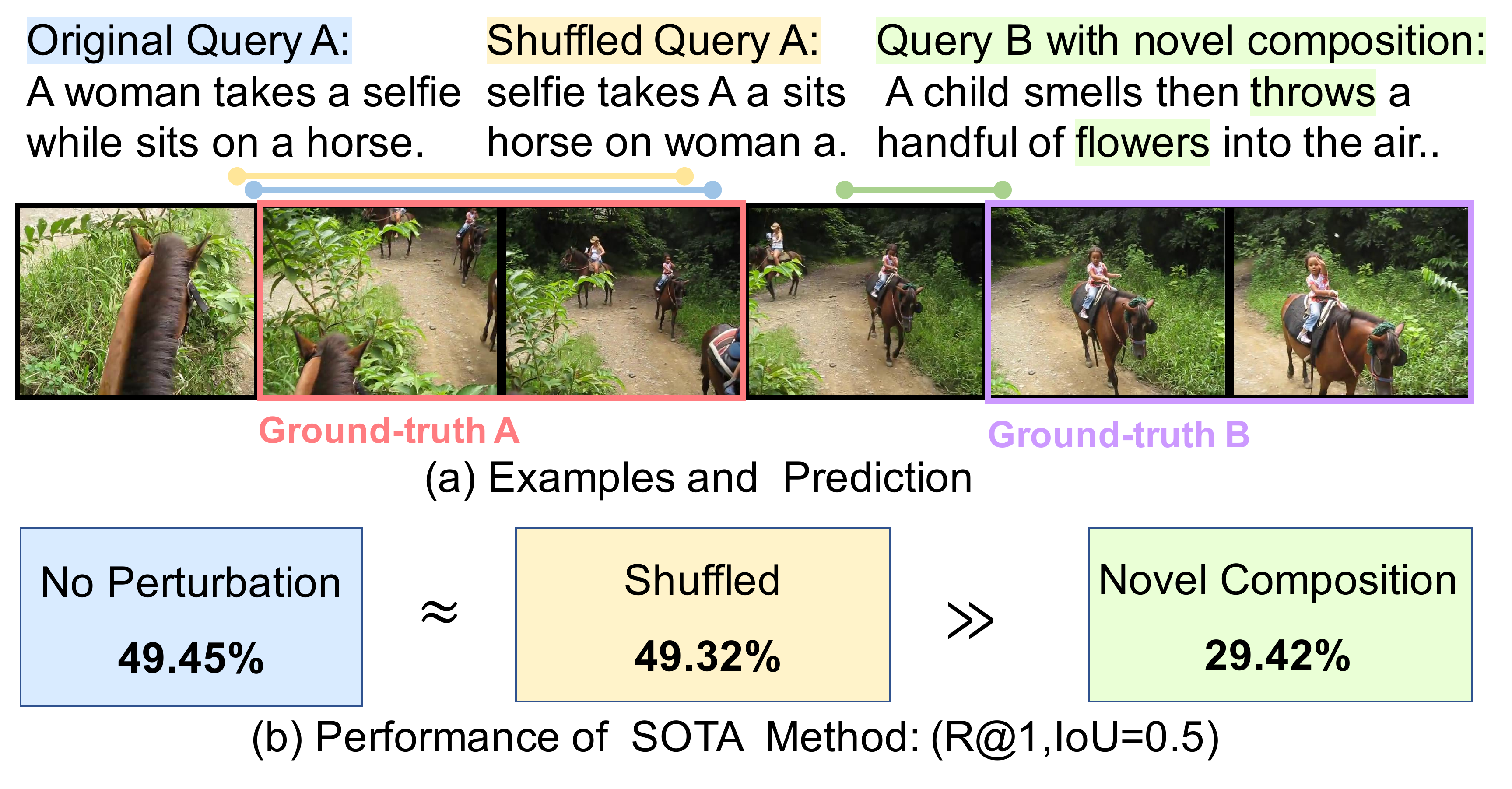}
		\vspace{-0.7cm}
		\caption{(a) On the top, we show three examples of two queries. (b) On the bottom, we report comparisons on Charades-CG with metric R@1, IoU@0.5. The left blue box represents the original model. The middle yellow box represents the model with shuffled queries as input. The right green box represents the performance on the queries that contain novel compositions.}
		\label{demo}
		\vspace{-0.6cm}
	\end{figure}
	
	Although the compositional generalization is a key property of human language that allows temporal grounding beyond pre-defined classes, current temporal grounding datasets do not specifically test for this ability. The training and testing splits of existing datasets contain almost the same compositions~(\eg \textsl{verb-noun pair, adjective-noun pair, etc}). Our statistical results show that only 1.37\% and 5.19\% of testing sentences contain novel compositions in the Charades-STA~\cite{gao2017tall} and ActivityNet Captions~\cite{krishna2017dense} datasets, respectively. To systematically measure the compositional generalizability~(CG) of existing methods, we introduce a new task, \textbf{Compositional Temporal Grounding}. Our compositional temporal grounding task aims to test whether the model can generalize to the sentences that contain novel compositions of seen words. We construct two re-organized datasets \textbf{Charades-CG} and \textbf{ActivityNet-CG}. Our dataset split protocols enable us to measure whether a model can generalize to novel compositions, of which the individual components have been observed during training but the combination is novel.
	
	
	Using our newly constructed datasets, we evaluate modern state-of-the-art~(SOTA) temporal grounding models, and empirically find that SOTA models fail to achieve compositional generalization, though they have achieved promising progress on the typical temporal grounding task. We observe that their performance drops dramatically~(Figure \ref{demo}.b, left vs. right). The results indicate that the SOTA models may not well generalize to novel compositions. Furthermore, as word order is a crucial factor for the compositionality of language, we analyze the word order sensitivity of SOTA models to gain more intuitive insight. Specifically, we randomly shuffled queries in advance and then use the shuffled sentences to train and evaluate the models. Surprisingly, we find that they are insensitive to the word order, even though permuting word order destroys the complete semantics of original sentences~(Figure \ref{demo}.b, left vs. middle). These observations confirm with recent studies~\cite{otani2020uncovering, yuan2021closer} suggesting that current models are heavily driven by superficial correlations. This pushes us to rethink the solution of temporal grounding.

	When we systematically analyze the SOTA models, we find that previous temporal grounding methods largely neglect the structured semantics in video and language, which is crucial for compositional reasoning. These methods~\cite{mun2020local, zhang2020learning, zhang2020span, gao2017tall} mainly encode both sentence and video segments into unstructured global representations, respectively, and then devise specific cross-modal interaction modules to fuse them for final prediction. These global representations fail to explicitly model video structure and language compositions. 
	Take the novel composition of ``throws flowers''  in Figure \ref{demo}.a as example. If the model infers the individual semantics of the two words, as well as establish the correspondence of them to specific semantics in video~(\ie \textsl{the action ``throw'' and the object ``flower'' in video }), the model can easily localize the novel composition in video by composing the corresponding video semantics of the two words. 
	
	Motivated by this insight, we propose a novel \textbf{V}ar\textbf{I}ational cro\textbf{S}s-graph re\textbf{A}soning~(\textbf{VISA}) framework for compositional temporal grounding. By explicitly modeling the semantic structures of video and language, and inferring the fine-grained correspondence between them, our VISA model can achieve joint compositional reasoning. Specifically, we first introduce a hierarchical semantic graph that explicitly decomposes both video and language into three semantic hierarchies~(\ie \textsl{global events, local actions, and atomic objects}). The hierarchical semantic graph serves as unified structured representations for both video and language, which tightly couple multi-granularity semantics between the two modalities. Second, we propose a variational cross-graph correspondence learning that establishes fine-grained semantic correspondence between the semantic hierarchical graphs of video and language. 
	
	Our contributions are summarized as follows:
	
	\vspace{-0.2cm}
	\begin{itemize}

		\item We introduce a new task, Compositional Temporal Grounding, as well as new splits of two prevailing temporal grounding datasets, which are able to measure the compositional generalizability of existing methods.
		
		\item We perform in-depth analyses on several SOTA models and empirically find that they fail to achieve compositional generalization

		\item We propose a \textbf{V}ar\textbf{I}ational cro\textbf{S}s-graph re\textbf{A}soning (\textbf{VISA}) framework that decomposes video and language into hierarchical graphs and learns fine-grained cross-graph correspondence between them.
		
		\item Experimental results validate the significant superiority of our approach on compositional generalizability.

	\end{itemize}
	
	\section{Related Work}
	
	\noindent
	\textbf{Temporal Grounding.} Recently, the development of deep learning~\cite{lecun2015deep, li2020ib} promotes the prosperity of computer vision \cite{zhang2022boostmis, li2020multi, guo2021semi} and vision-and-language understanding \cite{zhang2020relational, li2020unsupervised, li2019walking, Zhang_Jiang_Wang_Kuang_Zhao_Zhu_Yu_Yang_Wu_2020, Zhang_Tan_Zhao_Yu_Kuang_Jiang_Zhou_Yang_Wu_2020, Zhang_Yao_Zhao_Chua_Wu_2021}. Temporal grounding in videos via language is a recently proposed task~\cite{gao2017tall, krishna2017dense}. Existing supervised methods can be categorized into two groups. 1)~Proposal-based methods~\cite{gao2017tall, zhang2020learning, zhang2019man, yuan2019semantic } first extract candidate proposals by temporal sliding windows and then match the query sentence with them by multi-modality fusion. 
	2)~Proposal-free methods~\cite{mun2020local, zhang2020span, yuan2019find, Li2022End2End} directly predict the temporal boundaries of target segments without pre-defining proposals. 
	 In this paper, we evaluate the compositional generalizability of current methods.
	
	

	\noindent
	\textbf{Compositional Generalization.} Recently, compositional generalization has received increasing attention as its advantages on robustness and sample efficiency. To evaluate the compositional generalization, Lake~\textsl{et al.}~\cite{lake2018generalization} propose the SCAN benchmark, which requires translating instructions generated by a phrase-structure grammar to action sequences. The SCAN is split such that the testing set contains unseen compositions in the training set. The following works have proposed several techniques to improve SCAN, including data augmentation~\cite{andreas2019good}, meta-learning~\cite{lake2019compositional, nye2020learning, ding2022learning}, and architectural design~\cite{gordon2019permutation, ding2022nap}. Some recent works also explore compositional generalization on other applications, including image captioning~\cite{nikolaus2019compositional, zhang2021consensus, zhang2021magic}, visual question answering~\cite{grunde2021agqa, johnson2017clevr}, action recognition~\cite{materzynska2020something, zhukov2019cross, wray2019fine}, and state-object recognition~\cite{mancini2021open}. In this paper, we systematically study the compositional generalization on temporal grounding natural language sentences in videos.
	

	
	\noindent

	\section{Compositional Temporal Grounding}

	\vspace{-0.1cm}
	\subsection{Problem Formulation}
	\vspace{-0.2cm}
	
	To systematically benchmark the progress of current methods on compositional generalization, we introduce a new task, \textbf{Compositional Temporal Grounding}. Our compositional temporal grounding task aims to evaluate how well a model can generalize to query sentences that contain novel compositions or novel words. We construct new splits of two prevailing datasets Charades-STA~\cite{gao2017tall} and ActivityNet Captions~\cite{krishna2017dense}, named \textbf{Charades-CG} and \textbf{ActivityNet-CG}, respectively. Specifically, we define two new testing splits: Novel-Composition and Novel-Word. Each sentence in the novel-composition split contains one type of novel composition. We define the composition as novel composition if its constituents are both observed during training but their combination way is novel. Each sentence in the novel-word splits contains a novel word, which aims to test whether a model can infer the potential semantics of the unseen word based on the other learned composition components appearing in the context.


	\vspace{-0.1cm}
	\subsection{Dataset Re-splitting}
	
	For each dataset, we first combine all instances in the original training set and testing set, and remove the instances that can be easily predicted solely based on videos. We then re-split each dataset into four sets: training, novel-composition, novel-word, and test-trivial. The test-trivial set is similar to the original testing set, where most of the compositions are seen during training. Concretely, We use AllenNLP~\cite{gardner2018allennlp} to lemmatize and label all nouns, adjectives, verbs, adverbs, prepositions in language queries. Based on dependency parsing results, we define 5 types of compositions: verb-noun, adjective-noun, noun-noun, verb-adverb, and preposition-noun. For each type of composition, we construct a statistical table, where the row indexes are all possible first components of the composition and the column indexes are all possible second components of the composition. Taking verb-noun as an example, the element in row $i$ and column $j$ corresponds to the composition that consists of the $i-th$ verb and the $j-th$ noun in the dataset. For each table, we first sample an element from each row and each column, and then add all queries that contain the sampled compositions to the training set, which ensures that all components of compositions can be observed in the training set. Next, for each type of composition, we sample compositions from tables and add the corresponding queries into the novel-composition split. Meanwhile, we sample some words as new words and add the queries that contain the new words into the novel-word split. Since each video is associated with multiple text queries, if one query is selected into the training set, we will add other queries of the same video into the training set. If one query is selected into the novel-composition or novel-word split, we will add the remaining queries of the same video into the test-trivial set. Thus, we make sure that there is no video overlap between training and testing sets. Table \ref{t1} summarizes detailed statistics. We provide more details in supplementary materials.
	
	\begin{table}[!t]
		\resizebox{\linewidth}{!}{
			\centering
			\begin{threeparttable}
				\begin{tabular}{ llcc}
					\hline
					Dataset &Split  &Videos  &Queries\\		
					
					\hline    
					\multirow{4}*{Charades-CG}  &Training  &3555   &8281 \\
					&Novel-Composition  &2480      &3442\\
					&Novel-Word     &588        &703\\
					&Test-Trivial  &1689      &3096\\
					
					\hline
					\multirow{4}*{ActivityNet-CG}  &Training  &9659    &36724  \\
					&Novel-Composition       & 4202    &12028    \\
					&Novel-Word         &2011    &3944\\
					&Test-Trivial      &4775    &15712  \\
					
					\hline
				\end{tabular}
			\end{threeparttable}
		}
		\vspace{-0.3cm}
		\caption{Statistics of Charades-CG and Activity-CG.}
		\label{t1}
		\vspace{-0.5cm}
	\end{table}

	\begin{figure*}[!t]
		\centering
		\includegraphics[width=\textwidth]{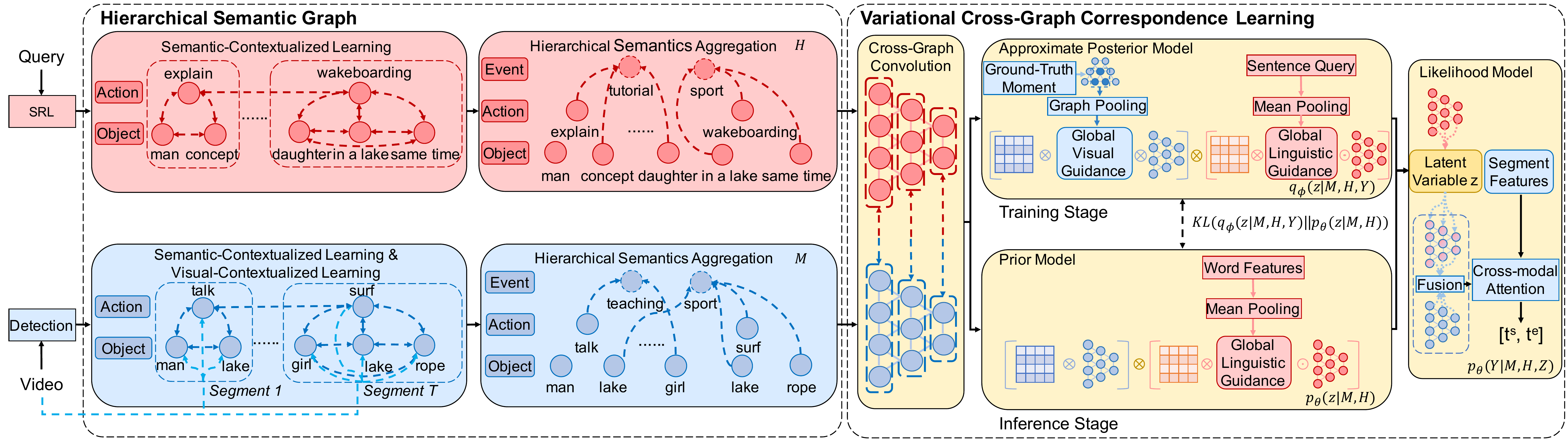}
		\vspace{-0.7cm}
		\caption{Overview of our VISA framework. We omit the details of the input video and sentence. 
		}
		\label{overview}
		\vspace{-0.55cm}
	\end{figure*}

	\section{Method}
	
	
	As illustrated in Figure \ref{overview}, our VISA framework mainly consists of two components: a \textbf{hierarchical semantic graph} and a \textbf{variational cross-graph correspondence learning}. Given an untrimmed video $V$ and a query sentence $Q$, the hierarchical semantic graph first decomposes them into three semantic hierarchies~(\ie \textsl{global events, local actions, and atomic objects}), respectively. Then, the variational cross-graph correspondence learning establishes fine-grained semantic correspondence between two graphs. Finally, based on the fine-grained semantic correspondence between video and sentence, our VISA infers the target moment that semantically corresponds to the given query.
	
	
	\vspace{-0.05cm}
	\subsection{Hierarchical Semantic Graph}\label{5.1}
	\vspace{-0.1cm}
	Language queries describe some semantic events~\cite{li2021adaptive}, which can be further parsed to central predicates and their corresponding arguments. Similarly, videos naturally record some relevant events in our lives, which consist of a variety of actions and each action involves multiple objects. Therefore, language and videos are both inherently organized in hierarchical structures. Based on this observation, given a video $V$ and a query $Q$, we decompose both of them into three semantic hierarchies, which correspond to global events, local actions, and atomic objects, respectively. Such a hierarchical semantic graph provides a unified structured representation for modeling fine-grained semantic correspondence between videos and language queries. 
	

	\noindent
	\textbf{Graph Initialization.} For an untrimmed video $V$, we first divide it into a sequence of segments with a fixed length and extract the features using pre-trained 3D CNN: $\{V_t\}_{t=1}^T$, where $V_t=\{f_i^t\}_{i=1}^K$ and $f_i^t$ denotes the C3D features of frame $i$ in segment $t$. Then, we adopt off-the-shelf object detection and action recognition models to extract objects and actions for each segment, where each segment contains $N_1$ object nodes $\{\overline{s}^{o}_{t, i}\}_{i=1}^{N_1} \in \mathbb{R}^{d \times N_1}$ and $N_2$ action nodes $\{\overline{s}^{a}_{t, i}\}_{i=1}^{N_2} \in \mathbb{R}^{d \times N_2}$. We initialize both object and action nodes by the sum of the GloVe~\cite{pennington2014glove} vectors of each word in the object labels and action labels. Finally, all the object nodes $\overline{S}^{o}$ and action nodes $\overline{S}^{a}$ across segments constitute the first and second hierarchies of the video semantic graph.
	
	For query $Q$, we use semantic role labeling~(SRL) to decompose the query into multiple semantic structures. Each semantic structure contains a central predicate~(verb) and some corresponding arguments~(noun phrases including prep, adj, and adv). The predicates are considered as action nodes denoted by $\{\overline{c}^{a}_{i}\}_{i=1}^{L_2} \in \mathbb{R}^{d \times L_2}$, and the arguments are considered as object nodes denoted by $\{\overline{c}^{o}_{i, j}\}_{j=1}^{L_1} \in \mathbb{R}^{d \times L_1}$. If a word serves as multiple arguments for different predicates, we duplicate it for each action node. Similarly, we initialize them using GloVe word vectors. Finally, all the object nodes $\overline{C}^o$ and action nodes $\overline{C}^a$ constitute the first and second hierarchies of the language semantic graph.
	

	\noindent
	\textbf{Semantic-Contextualized Learning.} Events are high-level semantic abstractions of video context and involve complicated interactions between different semantic concepts. For example, the query ``the camel stands up and walks off with the family riding on its back'' is composed of objects~(\emph{camel}, \emph{the family}), actions~(\emph{stands up, walks off, and riding on}), and the underlying relations among them such as the spatial relation~(\emph{on its back}), the temporal relation~(\emph{stands up and walks off}), and the agentive relation~(\emph{riding on}). Therefore, to achieve comprehensive understanding of video events, we present semantic-contextualized learning to model the complicated interaction between the semantic nodes and learn fine-grained contextual information beyond the coarse semantic labels. Further, semantic contextual information is crucial for resolving semantic ambiguity of individual semantic nodes as the pre-trained detector might be noisy and the detected actions and objects might have dramatic variations in appearance.
	
	Concretely, we define three types of undirect edges: \emph{action-action}, \emph{action-object}, and \emph{object-object}. For the video semantic graph (/the language semantic graph), the object nodes in the same segment (/semantic structure) are connected by the \emph{object-object} edges, the action and object nodes in the same segment (/semantic structure) are connected by the \emph{action-object} edges, and all the action nodes are connected by the  \emph{action-action} edges. Afterward, we perform relation-aware graph convolution on video semantic graph. For a semantic node $\overline{s}_i \in \{\overline{S}^a, \overline{S}^o\}$, we calculate the adjacency correlation for each edge type $r$ as:
	
	\begin{equation}
	\tilde{\alpha}_{ij}^r = (W^r \overline{s}_i)^T\cdot(W^r \overline{s}_j), \quad \alpha_{ij}^r  = \frac{exp(\tilde{\alpha}_{ij}^r)}{\sum_{j \in \mathcal{N}_i^r} exp(\tilde{\alpha}_{ij}^r)}
	\end{equation}
	
	\noindent
	where $\mathcal{N}_i^r$ is the neighborhood nodes of $s_i$ on edge type $r$ and $W_r$ is the relation-specific projection matrix. Then, we refine $s_i$ using the neighboring nodes of all edge types as: 
	
	\begin{equation}
	\hat{s}_i = \sum_{r \in R} \sum_{j \in \mathcal{N}_i^r} \alpha_{ij}^r \cdot (U^r \overline{s}_j)
	\end{equation}
	
	\noindent
	where $R$ is the three types of edges and $U_r$ is another transformation matrix. $\hat{s}_i$ is the result of the first relation-aware graph convolution layer. To model multi-order relations, we perform $M$ layers of relation-aware graph convolution and learn final semantic-contextualized node features $S=\{s_i\}_{i=1}^{N_v} \in \mathbb{R}^{N_v \times d}$, where $N_v$ is the total number of action and object nodes. In the same manner, we can obtain semantic-contextualized node features $C=\{c_i\}_{i=1}^{N_s} \in \mathbb{R}^{N_s \times d}$ of language semantic graph.
	
	\noindent
	\textbf{Visual-Contextualized Learning.} We further propose visual-contextualized learning to collect relevant visual context from videos to the video semantic graph. Specifically, for a semantic node $s_i$, let $V_i = \{f_j^i\}_{j=1}^K$ denotes the corresponding segment, and $f_j^i$ is the frame feature~(following, we omit the superscript i for simplicity). We first compute the visual filter for each frame $f_j$ in the segment and obtain the filtered visual feature as:
	
	\begin{equation}
	g^i_j = \sigma(W^g[s_i; \overline{f}; f_j] + b_g), \quad f_j^{'} = f_j \odot g^i_j
	\end{equation}
	
	\noindent
	where $\odot$ denotes the Hadamard product, and $\overline{f}$ is obtained by performing average pooling on the $V_i$. Then, we perform max-pooling across the filtered frame features to get the semantic-relevant visual context as $F_i = MaxPool(f_1^{'}, ..., f_K^{'})$. Finally, we concatenate $s_i$ with $F_i$ and transform them to the original dimension by a transformation matrix $W_v$ as $s_i = W^v[s_i, F_i]$. Here, we reuse $s_i$ to represent the final visual-contextualized semantic node representation for simplicity.

	\noindent
	\textbf{Hierarchical Semantics Aggregation.} Based on the observation that semantic events are composed of a series of interactional actions and objects, we propose the hierarchical semantic aggregation mechanism, which aggregates the semantics from the contextualized action nodes and object nodes to compose the global event nodes. Inspired by the success of positional query encoding~\cite{carion2020end} in object detection, we initialize the event nodes as a set of learnable query vectors $\{p_i\}_{i=1}^{N_p}$ and then aggregate relevant semantics from action nodes and object nodes to refine the event nodes. Here we take the video semantic graph as an illustration. For an event query $p_i$, we calculate the attention weights over semantic nodes $\{s_j\}_{j=1}^{N_v}$ and update the $p_i$, given by:
	
	\begin{equation}
	\tilde{p_i}\!=\!\sum_{j=1}^{N_v} \alpha_{ij}^{e}\cdot s_j, \alpha_{ij}^{e}\!=\!\frac{exp({(W^e_1 p_i)}^T \cdot (W^e_2 s_j))}{\sum_{j=1}^{N_v} exp({(W^e_1 p_i)}^T \cdot (W^e_2 s_j))}
	\end{equation}
	
	\noindent
	where $W^e_1, W^e_2$ are projection matrices, and $\tilde{p_i}$ is the semantics-aware event node. Subsequently, we stack multiple such graph self-attention layers and merge the final event nodes into $\{s_j\}_{j=1}^{N_v}$ to form the complete hierarchical semantic graph of video, denoted by $M = \{m_i\}_{i=1}^{N_m} \in \mathbb{R}^{d \times N_m}$. In the same manner, we can obtain the complete hierarchical semantic graph of language, denoted by $H = \{h_i\}_{i=1}^{N_h} \in \mathbb{R}^{d \times N_h}$. $M$ and $H$ are the unified structure of three semantic hierarchies, which tightly couple multi-granularity semantics between the two modalities.

	\subsection{Variational Cross-Graph Correspondence}\label{5.2}
	After parsing both videos and language queries into individual hierarchical semantic graphs, we then model the cross-modality interactions between two graphs by cross-graph convolution, and induce the fine-grained semantic correspondence between them for final prediction. The objective function can be formulated as $P(Y|M, H)$, where $Y$ is the target time interval. Since the ground-truth correspondence between two graphs is not available, we treat the cross-graph correspondence as a latent variable $z$. The problem can then be formulated into a variational inference framework~\cite{sohn2015learning} and the objective function can be rewritten as $P(Y|M, H, z)P(z|M, H)$. Instead of directly maximizing $P(Y|M, H)$, we propose to maximize its evidence lower bound~(ELBO)~\cite{kingma2013auto} as follows:
	
	\begin{equation}\label{e5}
	\begin{aligned}
	\mathcal{L}^{ELBO}(\phi, \theta) \quad = & \quad E_{q_{\phi}(z|M, H, Y)}log p_{\theta}(Y|M, H, z) \\
	&- KL(q_{\phi}(z|M, H, Y) || p_{\theta}(z|M, H)) \\
	&\leq log p(Y|M, H)
	\end{aligned}
	\end{equation}
	\noindent
	Specifically, we characterize $P(Y|M, H)$ using three components: a prior model $p_{\theta}(z|M, H)$, a posterior model $q_{\phi}(z|M, H, Y)$, and a likelihood model $p_{\theta}(Y|M, H, z)$. In the following, we first introduce cross-graph convolution to capture the semantic correlation between two graphs and then describe these three models in detail.
	
	\noindent
	\textbf{Cross-Graph Convolution.} Given the graphs $M$ and $H$, we perform cross-graph convolution between the same hierarchical levels of two graphs. For a video semantic node $m_i^k$, the cross-convolution from $H$ to $M$ is formulated as:
	
	\begin{equation}
	\alpha_{ij}^{h2m} = \frac{exp({(W^c_1 m_i^k)}^T \cdot (W^c_2 h_j^k))}{\sum_{j \in \mathcal{N}_H^k} exp({(W^c_1 m_i^k)}^T \cdot (W^c_2 h_j^k))}
	\end{equation}
	
	\begin{equation}
	\tilde{m}_i^k = (1 - \beta_i^k)\odot m_i^k  +  \beta_i^k \odot \sum_{j \in \mathcal{N}_H^k} \alpha_{ij}^{h2m} \cdot h_j^k, k \in \{e, a, o\}
	\end{equation}
	
	\noindent
	where $\beta_i^k\!=\!\sigma(U^gm_i^k + b)$ controls the information flow from $H$ to $M$, $k$ denotes three semantic levels (\ie \textsl{event, action, object}), $\mathcal{N}_H^k$ denotes the nodes of $H$ in level $k$. In a similar manner but reversed order, we can obtain $\tilde{H}$.
	
	\noindent
	\textbf{Prior Model.} Given $\tilde{M}$ and $\tilde{H}$, the prior model $p_{\theta}(z|M, H)$ aims to infer the cross-graph correspondence captured by a latent variable $z \in \mathbb{R}^{N_m \times N_h}$, where $z_{ij}$ corresponds to the semantic correspondence between $\tilde{m}_i$ and $\tilde{h}_j$. 
	Specifically, the $z_{ij}$ can be formulated as:
	
	\begin{equation}
	\tilde{z}_{ij} = {(W^s_1 \tilde{m}_i)}^T \cdot (W^s_2 q \odot \tilde{h_j}), z_{ij} = \frac{exp(\tilde{z}_{ij})}{\sum_{j=1}^{N_h} exp(\tilde{z}_{ij})}
	\end{equation}
	
	\noindent
	where $q$ is the global sentence feature that guides the semantic correspondence inference.

	\noindent
	\textbf{Approximate Posterior Model.} The posterior model $q_{\phi}(z|M, H, Y)$ infers the cross-graph correspondence with additional information of ground-truth $Y$. According to the temporal boundary $Y$, we can determine the segments in $Y$ and the action and object nodes that correspond to these segments. These nodes in the video graph contain the most relevant semantics to the language semantic graph, which can better guide cross-graph correspondence learning. Therefore, we obtain $m^*$ through mean-pooling over these nodes and use $m^*$ to guide the correspondence learning:
	
	\begin{equation}
	\tilde{z}_{ij} = {(W^s_3 m^* \odot \tilde{m}_i)}^T \cdot (W^s_4 q \odot \tilde{h_j}),\ z_{ij} = \frac{exp(\tilde{z}_{ij})}{\sum_{j = 1}^{N_h} exp(\tilde{z}_{ij})}
	\end{equation}
	
	\noindent
	where $m^*$ and $q$ serve as global visual and linguistic guidance, respectively.

	\noindent
	\textbf{Likelihood Model.} The likelihood model $p_{\theta}(Y|M, H, z)$ predicts the temporal boundary based on the latent correspondence $z$ and hierarchical semantic graphs $\tilde{M}$ and $\tilde{H}$. Specifically, we first integrate two graphs based on the learned cross-graph correspondence to obtain joint multi-modality representations:
	
	\begin{equation}
	M^{'} = z \tilde{H} \in \mathbb{R}^{d \times N_m}, \quad M^J = W^J [\tilde{M}; M^{'}] \in \mathbb{R}^{d \times N_m}
	\end{equation}
	
	\noindent
	where projection matrix $W^J\!\in\!\mathbb{R}^{d \times 2d}$ and $M^J$ is the joint multi-modality representations of the hierarchical semantic graph. Next, we use $M^J$ to refine segment features $X = \{x_t\}_{t=1}^T \in \mathbb{R}^{d \times T}$. We perform mean-pooling over frame features $\{f_i^t\}_{i=1}^K$ of segment $V_t$ to obtain the segment features $x_t$. We adopt multi-head cross-modal attention to softly select relevant information from $M^J$ to $X$. Concretely, we take $X$ as queries and $M^J$ as keys and values:
	
	\begin{equation}
	X^* = MultiAttn(X, M^J, M^J)
	\end{equation}
	
	\noindent
	where $X^*$ is the semantics-aware segment representations. Subsequently, we summarize the segment representations using attentive pooling based on the sentence feature $q$:
	
	\vspace{-0.5cm}
	\begin{equation}
	v^* = \sum_{i=1}^T \alpha^q_{i} \cdot x^*_i, \quad\alpha^q_{i} = \frac{exp((W^q_1q)^T\cdot(W^q_2x_i^*))}{\sum_{i=1}^T exp((W^q_1q)^T\cdot(W^q_2x_i))}
	\end{equation}
	
	\noindent
	where $v^*$ is the summarized video feature. Finally, we predict the time interval $(t^s, t^e)$ as $t^s, t^e = MLP(v^*)$.
	
	\subsection{Optimization}
	As described in Equation \ref{e5}, the ELBO objective function consists of two terms. The first term corresponds to the negative number of the regression loss. Specifically, following \cite{mun2020local}, we minimize the sum of smooth $L_1$ distances between the normalized ground-truth time interval $(\hat{t^s}, \hat{t^e}) \in [0, 1]$ and our prediction $(t^s, t^e)$. This term not only teaches the likelihood model to predict the correct time interval but also encourages the approximate posterior model to learn more accurate cross-graph correspondence. The second term corresponds to the KL-divergence loss. Concretely, as the latent variable $z$ is a correlation matrix, we compute the KL-divergence by rows. Intuitively, through minimizing this term, we can teach the prior model to capture the cross-graph semantic correspondence as well as the approximate posterior model. During testing without access to the ground-truth, we can use the learned prior model to replace the approximate posterior model to infer the cross-graph correspondence. Note that we use the approximate posterior model to generate $z$ during training.

	\linespread{0.93}
	\begin{table*}[!htb]
		\resizebox{\textwidth}{!}{
			\centering
			\begin{tabular}{ lll ccc ccc ccc}
				\hline
				&\multicolumn{2}{l}{\multirow{2}*{Method}} 
				&\multicolumn{3}{c}{\textsl{Test-Trivial}} 
				&\multicolumn{3}{c}{\textsl{Novel-Composition}}
				&\multicolumn{3}{c}{\textsl{Novel-Word}}
				\\
				
				\cmidrule(lr){4-6} \cmidrule(lr){7-9}  \cmidrule(lr){10-12} 
				&\multicolumn{2}{l}{} &IoU=0.5 &IoU=0.7 &\multicolumn{1}{c}{mIoU}  &IoU=0.5 &IoU=0.7 &\multicolumn{1}{c}{mIoU} &IoU=0.5 &IoU=0.7 &\multicolumn{1}{c}{mIoU}  \\
				\hline

				\multicolumn{1}{l}{\multirow{1}*{Weakly-supervised}} &\multicolumn{2}{l}{WSSL}  
				&15.33  &5.46    &18.31    &3.61 &1.21 &8.26  &2.79  &0.73  &7.92  \\
				\hline
				
				\multicolumn{1}{l}{\multirow{1}*{RL-based}} &\multicolumn{2}{l}{TSP-PRL}  
				&39.86 & 21.07   &38.41    &16.30  &2.04 &13.52   &14.83  &2.61  &14.03   \\
				\hline
				
				\multicolumn{1}{l}{\multirow{2}*{Proposal-based}} &\multicolumn{2}{l}{TMN} 
				&18.75    &8.16  &19.82  &8.68    &4.07  &10.14  &9.43    &4.96  &11.23 \\

				\multicolumn{1}{l}{}&\multicolumn{2}{l}{2D-TAN} 
				&48.58 &26.49  &44.27 &30.91  &12.23  &29.75 &29.36  &13.21  &28.47 \\
				
				\hline
				
				\multicolumn{1}{l}{\multirow{3}*{Proposal-free}} &\multicolumn{2}{l}{LGI} 
				&49.45  &23.80    &45.01  &29.42 &12.73   &30.09  &26.48  &12.47 &27.62\\

				\multicolumn{1}{l}{}&\multicolumn{2}{l}{VLSNet}  
				&45.91  &19.80   &41.63   &24.25  &11.54 &31.43 &25.60  &10.07 &30.21  \\

				\multicolumn{1}{l}{} &\multicolumn{2}{l}{\textbf{Ours-VISA}}  
				&\textbf{53.20}  &\textbf{26.52}    &\textbf{47.11}  &\textbf{45.41} &\textbf{22.71} &\textbf{42.03} &\textbf{42.35} &\textbf{20.88} &\textbf{40.18}  \\
				\hline

			\end{tabular}
			
		}
		\vspace{-0.3cm}
		\caption{Performances (\%) of SOTA temporal grounding models and our VISA on the proposed Charades-CG datasets.}
		\label{t2}
		\vspace{-0.2cm}
	\end{table*}

	\begin{table*}[!htb]
		\resizebox{\textwidth}{!}{
			\centering
			\begin{tabular}{ lll ccc ccc ccc}
				\hline
				&\multicolumn{2}{l}{\multirow{2}*{Method}} 
				&\multicolumn{3}{c}{\textsl{Test-Trivial}} 
				&\multicolumn{3}{c}{\textsl{Novel-Composition}}
				&\multicolumn{3}{c}{\textsl{Novel-Word}}
				\\
				
				\cmidrule(lr){4-6} \cmidrule(lr){7-9}  \cmidrule(lr){10-12} 
				&\multicolumn{2}{l}{} &IoU=0.5 &IoU=0.7 &\multicolumn{1}{c}{mIoU}  &IoU=0.5 &IoU=0.7 &\multicolumn{1}{c}{mIoU} &IoU=0.5 &IoU=0.7 &\multicolumn{1}{c}{mIoU}  \\
				\hline

				\multicolumn{1}{l}{\multirow{1}*{Weakly-supervised}} &\multicolumn{2}{l}{WSSL}  
				&11.03 &4.14    &15.07 &2.89  &0.76 &7.65 &3.09 &1.13 &7.10 \\
				\hline
				
				\multicolumn{1}{l}{\multirow{1}*{RL-based}} &\multicolumn{2}{l}{TSP-PRL}  
				&34.27   &18.80    &37.05 &14.74 &1.43  &12.61 &18.05 &3.15 &14.34 \\
				\hline
				
				\multicolumn{1}{l}{\multirow{2}*{Proposal-based}} &\multicolumn{2}{l}{TMN} 
				&16.82   &7.01 &17.13 &8.74   &4.39 &10.08  &9.93   & 5.12 &11.38 \\

				\multicolumn{1}{l}{}&\multicolumn{2}{l}{2D-TAN} 
				&44.50	&26.03	&42.12 &22.80  &9.95  &28.49 &23.86 &10.37 &28.88 \\
				
				\hline
				
				\multicolumn{1}{l}{\multirow{3}*{Proposal-free}} &\multicolumn{2}{l}{LGI} 
				&43.56 &23.29  &41.37 &23.21  &9.02  &27.86 &23.10 &9.03 &26.95\\

				\multicolumn{1}{l}{}&\multicolumn{2}{l}{VLSNet}  
				&39.27   &23.12     &42.51  &20.21 &9.18 &29.07 &21.68 &9.94 &29.58\\

				\multicolumn{1}{l}{} &\multicolumn{2}{l}{\textbf{Ours-VISA}}  
				&\textbf{47.13}  &\textbf{29.64}    &\textbf{44.02}  &\textbf{31.51} &\textbf{16.73} &\textbf{35.85} &\textbf{30.14} &\textbf{15.90} &\textbf{35.13}  \\
				\hline

			\end{tabular}
			
		}
		\vspace{-0.3cm}
		\caption{Performances (\%) of SOTA temporal grounding models and our VISA on the proposed ActivityNet-CG datasets.}
		\label{t3}
		\vspace{-0.45cm}
	\end{table*}

	\vspace{-0.1cm}
	\section{Experiments}
	
	\vspace{-0.1cm}
	\subsection{Benchmarking the SOTA Methods}
	We evaluate the compositional generalizability of SOTA methods on the proposed Charades-CG and ActivityNet-CG datasets. Specifically, these methods can be categorized into four groups: 1) Proposal-based methods: \textbf{TMN}~\cite{liu2018temporal}, \textbf{2D-TAN}~\cite{zhang2020learning}; 2) Proposal-free methods: \textbf{LGI}~\cite{mun2020local}, \textbf{VLSNet}~\cite{zhang2020span}; 3) RL-based method: \textbf{TSP-PRL}~\cite{wu2020tree}; 4) Weakly-supervised method: \textbf{WSSL}~\cite{duan2018weakly}. Due to the space limitation, we provide more experimental results and implementation details in supplemental materials.
	
	\noindent
	\textbf{Evaluation Metrics.} Following previous works, we adopt ``R@n, IoU=m''  and mIoU~(\ie the average temporal IoU) as our evaluation metrics. Specifically, given a testing query, it first calculates the Intersection-over-Union (IoU) between the predicted moment and the ground truth, and ``R@n, IoU=m'' is defined as the percentage of at least one of top-n predictions with IoU larger than m.
	

	\vspace{-0.1cm}
	\subsection{Results on Compositional Temporal Grounding}
	
	Table \ref{t2} and Table \ref{t3} summarize the results of the above methods on compositional temporal grounding. Overall, our VISA achieves the highest performance on all dataset splits, demonstrating the superiority of our proposed model. Notably, we observe that the performance of all tested SOTA models drops significantly on the novel-composition and novel-word splits. The difference in performance between test-trivial and novel-composition~(novel-word) ranges up to 20\%. In contrast, our VISA surpasses them by a large margin on novel-composition and novel-word splits, demonstrating superior compositional generalizability. Particularly, for the novel-composition splits of Charades-CG and ActivityNet-CG datasets, our method significantly surpasses the SOTA methods by 30.86\% and 23.32\%  relatively on mIoU, respectively.

	\vspace{-0.1cm}
	\subsection{In-Depth Analysis}
	\noindent
	\textbf{Effect of Individual Components.} We conduct an ablation study to illustrate the effect of each component in Table \ref{t4}. Specifically, we train the following ablation models.
	 1) w/o SCL: we remove the Semantic-Contextualized Learning~(SCL). 2) w/o VCL: we remove the Visual-Contextualized Learning~(VCL). 3) w/o HSA: we remove the Hierarchical Semantics Aggregation~(HSA). 4) w/o VCC: we replace the Variational Cross-graph Correspondence learning~(VCC) by directly using cross-modal self-attention to fuse two graphs. 5) Detection-based: we directly use the detection results and SRL labels as features.

	The results of Row 1 and Row 2 indicate that learning fine-grained contextualized information is crucial for compositional reasoning. Also, the results of Row 3 validate the importance of event-level hierarchy on global semantic understanding. Ours w/o VCC does not achieve satisfying results, because directly fusing the graphs of video and sentence could possibly disrupt the semantic correspondence between them, which causes a negative effect on temporal grounding performance. In contrast, the proposed VCC establishes fine-grained cross-graph correspondence by variational inference, which is meticulous and achieves the best results. Furthermore, Row 5 suggests that the main performance gain does not directly come from the pre-trained detection models. Instead, these detected semantic labels serve as unified symbols for joint compositional reasoning.
	
	\begin{table}[!t]
		\centering
		\begin{tabular}{ lll|cc|cc}
			\hline
			&\multicolumn{2}{l|}{\multirow{2}*{Method}} 
			&\multicolumn{2}{c|}{Charades-CG} 
			&\multicolumn{2}{c}{ActivityNet-CG}
			\\
			
			&\multicolumn{2}{l|}{}  &Comp &\multicolumn{1}{c|}{Word}   &Comp &Word  \\
			\hline    
			1 &\multicolumn{2}{l|}{w/o SCL}   &43.75 &40.16       &29.03 &29.41                   \\
			2 &\multicolumn{2}{l|}{w/o VCL}   &42.26 &38.62      &29.34 &28.09					\\
			3 &\multicolumn{2}{l|}{w/o HSA}  &44.22 &41.09    &30.29	&29.31	\\
			4 &\multicolumn{2}{l|}{w/o VCC}  &41.08  &37.54   &27.32 &26.37 \\
			5 &\multicolumn{2}{l|}{Detection} &12.97  &11.70  &10.92 &10.07					\\
			\hline
			6 &\multicolumn{2}{l|}{{\small \textbf{VISA}}} &\textbf{45.41} &\textbf{42.35} &\textbf{31.51} &\textbf{30.14}					\\
			\hline
		\end{tabular}
		
		\vspace{-0.3cm}
		\caption{Ablation results with metric R@1, IoU=0.5 on novel-composition~(Comp) and novel-word~(Word) splits.}
		\label{t4}
		\vspace{-0.6cm}
	\end{table}

	\noindent
	\textbf{Results on Different Composition Types.} To gain further insight, we examine the results (R@1, IoU=0.5) of our models on different types of compositions. Table \ref{t5} shows that generalizing to ``Verb-Noun'' compositions is the most difficult, as it requires the model to accurately identify the corresponding action and objects in video and jointly reason over them to infer the semantics of the novel composition.
	

	\begin{table}[!t]
		\setlength\tabcolsep{2.5pt}
		\resizebox{\linewidth}{!}{
			\centering
			\begin{tabular}{ ll|ccc|ccc}
				\hline
				\multicolumn{2}{l|}{\multirow{2}*{Type}} 
				&\multicolumn{3}{c|}{Charades-CG} 
				&\multicolumn{3}{c}{ActivityNet-CG}
				
				\\
				\multicolumn{2}{l|}{}  &w/o VCC &w/o SCL  &\multicolumn{1}{c|}{VISA}  &w/o VCC &w/o SCL  &\multicolumn{1}{c}{VISA}   \\
				\hline    
				\multicolumn{2}{l|}{Verb-Noun}    &36.56    &38.82     &41.37    &24.41   &26.32   &28.89   \\
				\multicolumn{2}{l|}{Adj-Noun}     &42.17     &44.04    &45.06   &26.76     &28.31   &30.67	  \\
				\multicolumn{2}{l|}{Noun-Noun}  &40.38    &42.56     &43.41    &29.51  &30.20     &33.93     \\
				\multicolumn{2}{l|}{Verb-Adv}      &43.81    &46.37     &47.83   &31.08   &33.46    &35.60     \\
				\multicolumn{2}{l|}{Prep-Noun}    &44.12    &47.86     &48.61   &34.78   &36.03    &37.35   \\
				\hline
				
			\end{tabular}
		}
		\vspace{-0.3cm}
		\caption{Performance of our models on each composition type.}
		\label{t5}
		\vspace{-0.65cm}
	\end{table}	

	\noindent
	\textbf{Word Order Sensitivity.} To gain more intuitive insight, we explore whether the models are sensitive to the word order, which is a crucial factor for the compositionality of language. Intuitively, if we change the word order of a sentence, its semantics might change greatly and thus the original ground-truth temporal boundaries might not be suitable for the shuffled sentence. Specifically, we randomly shuffle queries in advance and then use the shuffled queries to train and evaluate the models. We define the sensitivity metric as the relative performance degradation of the shuffled version on R@1, IoU=0.5. The higher value indicates a higher sensitivity. According to Table~\ref{t6}, we surprisingly find that SOTA models are insensitive to the word order. In contrast, our method are sensitive to the linguistic structure of sentences. Moreover, we observe the highest sensitivity of our VISA on novel-word splits, indicating that the linguistic structure is important for inferring the semantics of novel words. In the end, we observe that the proposed SCL and VCC promote our method to capture the linguistic structure of sentences in a mutually rewarding way.


	\begin{table}[!t]
		\resizebox{\linewidth}{!}{
			\centering
			\begin{tabular}{l ccc ccc}
				\hline
				\multirow{2}*{Method}
				&\multicolumn{3}{c}{Charades-CG} 
				&\multicolumn{3}{c}{ActivityNet-CG}
				\\
				
				\cmidrule(lr){2-4} \cmidrule(lr){5-7} 
				&Trivial &Comp &\multicolumn{1}{c}{Word}   &Trivial &Comp &Word  \\
				\hline    
				2D-TAN       &0.41   &0.52   &0.43   &0.29   &0.30  &0.41 \\
				LGI              &0.28   &0.23   &0.16   &0.31   &0.22  &0.19 \\
				VLSNet       &0.07   &0.24   &0.10   &0.24   &0.31  &0.48 \\
				\hline
				VISA            &24.14  &29.80     &33.97   &22.09  &27.60   &31.89  \\
				w/o SCL  &19.64   &24.31   &29.72   &18.07    &24.64 &28.73  \\
				w/o VCC  &21.32   &26.73   &30.88   &20.15   &25.46  &29.79  \\
				\hline
			\end{tabular}
		}
		\vspace{-0.3cm}
		\caption{The word order sensitivity of SOTA models and VISA.}
		\label{t6}
		\vspace{-0.3cm}
	\end{table}
	
%
%

	\begin{figure}[h]
		\centering
		\includegraphics[width=\linewidth]{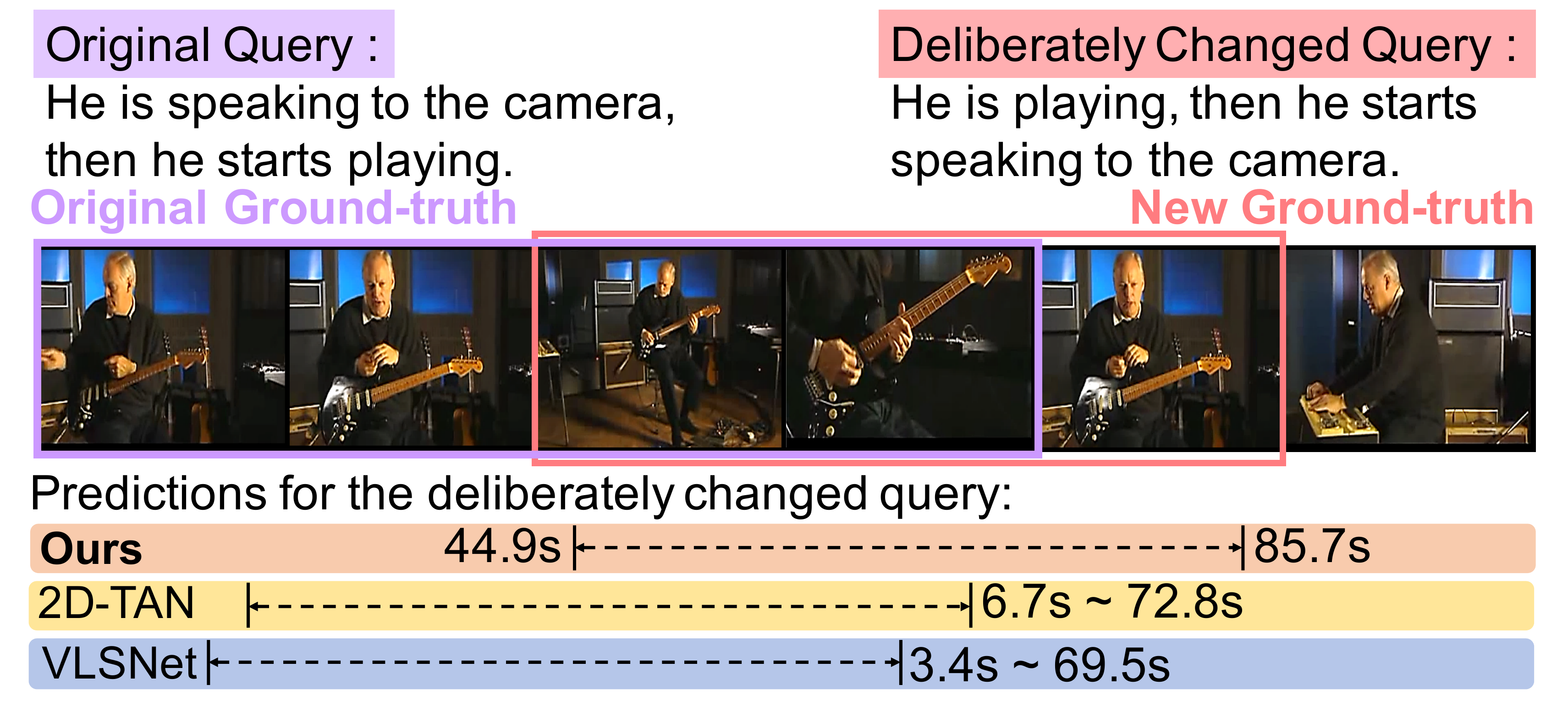}
		\vspace{-0.7cm}
		\caption{Qualitative examples on specifically shuffled queries.}
			
			
		\label{shuffle}
		\vspace{-0.5cm}
	\end{figure}
	
	\subsection{Qualitative Analysis}
	
	\noindent
	\textbf{Sensitivity on Specific Shuffling.} We manually select some query sentences and change their word order in some specific ways, such that the changed query can still semantically correspond to other segments in their original videos. As shown in Figure \ref{shuffle}, we annotate the changed query with a new ground-truth~(red box) and use the changed query to test models. Interestingly, the predictions of SOTA methods have higher IoU with the original temporal boundary, though the semantics of the sentence has been deliberately modified. In contrast, our VISA keenly captures the semantics change and locates to the new temporal boundary.
	
\begin{figure}[!t]
		\centering
		\includegraphics[width=\linewidth]{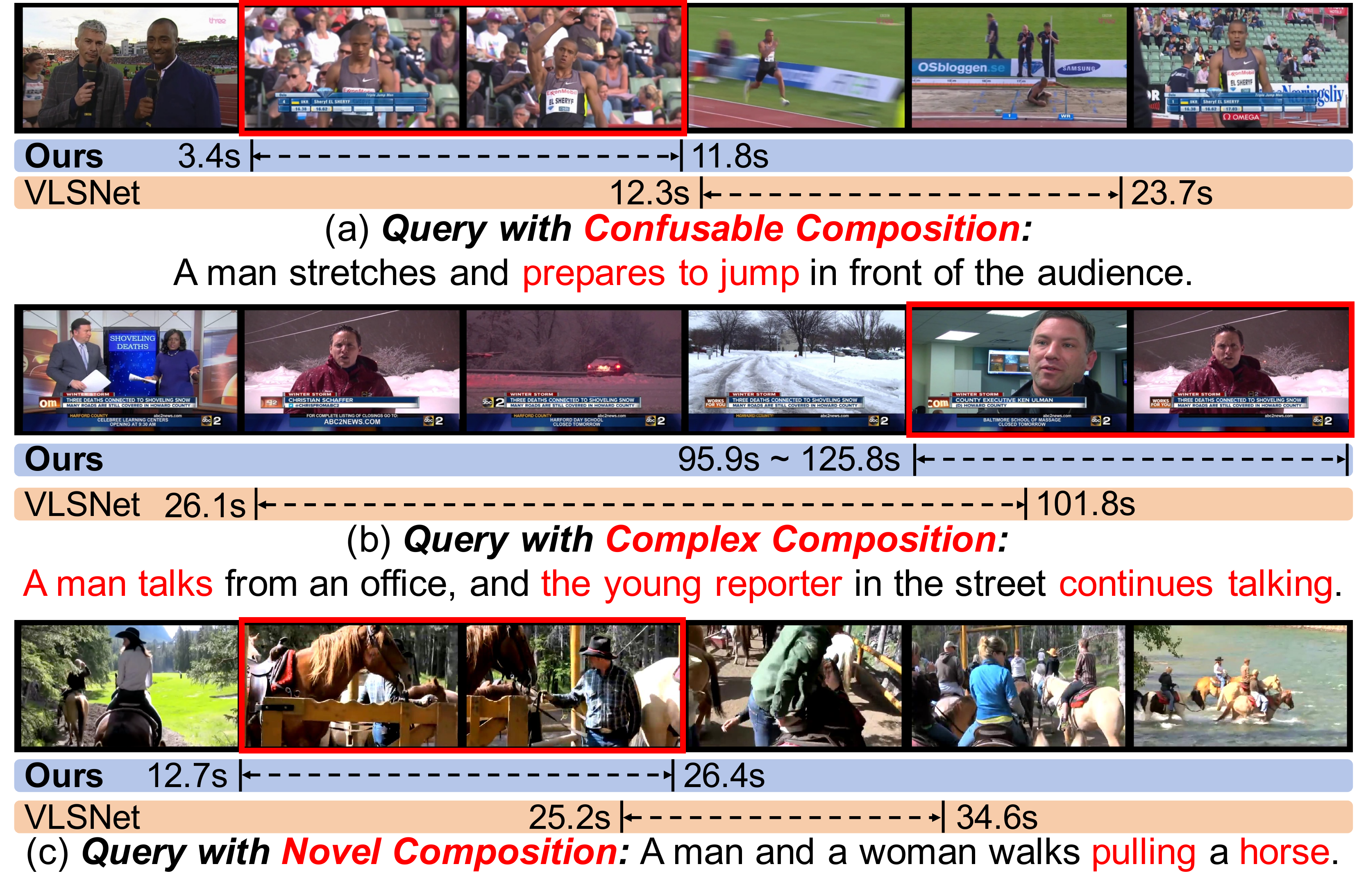}
		\vspace{-0.7cm}
		\caption{Qualitative examples of our VISA and VLSNet. The red boxes represent the ground-truth.
		}
		\label{case}
		\vspace{-0.3cm}
	\end{figure}

	\noindent
	\textbf{Qualitative Examples.} Figure \ref{case} visualizes three qualitative examples, which indicate the importance of compositionality. In the first case, the baseline fails to understand the composition meaning of ``prepares to jump '', so it mistakenly localizes to the ``jump'' segment. In contrast, our VISA successfully captures the compositional meanings. The second case contains complex compositions, which describe two events. Without inferring their temporal relationship from the composition structure, the baseline localizes a wrong segment, even though it also contains the two individual events (\ie \textsl{``a man talk''} and \textsl{``the reporter in the street talks''}). Conversely, our VISA understands the correct temporal order of these two events. The third case shows that our VISA successfully generalizes to novel composition. While \textsl{pulling}~(\eg \textsl{pulling rope}) and \textsl{horse}~(\eg \textsl{lead horse}) are both observed in the training split, the baseline suffers from generalizing to this novel composition.
	
\begin{figure}[!t]
		\centering
		\includegraphics[width=\linewidth]{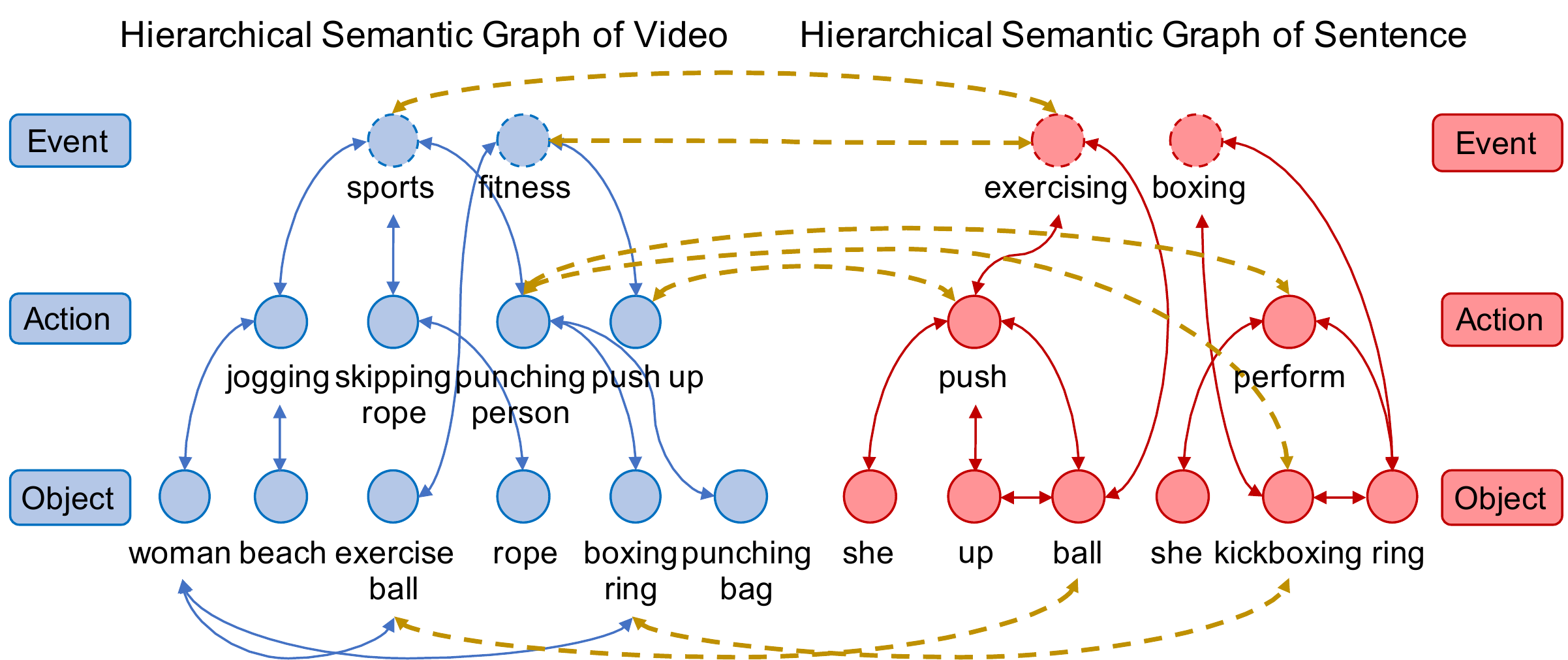}
		\vspace{-0.5cm}
		\caption{Visualization of the learned hierarchical semantic graph.
		}
		\label{vis}
		\vspace{-0.6cm}
	\end{figure}

	\noindent
	\textbf{Visualizing Learned Graph.} In Figure \ref{vis}, we present the learned hierarchical semantic graph. We visualize some key nodes and the edges with high weights. The yellow dotted lines represent the cross-graph semantic correspondence. If the semantic correspondence score between two nodes is greater than a specific threshold, we connect them with a yellow dotted line. We represent the event nodes by their most related semantics according to their attended nodes. Our VISA successfully aligns the visual semantics ``punching person'' and ``boxing ring' to the linguistic words ``perform kickboxing''.  Also, our VISA can connect ``push up'' and ``exercise ball'' to the words ``push up (with) ball''.

	\section{Conclusions}
	In this paper,  we introduce a new task, compositional temporal grounding to systematically evaluate the compositional generalizability of temporal grounding models. We conduct in-depth analyses on SOTA methods, and find they lack of compositional generalizability. We then introduce a novel VISA framework that learns fine-grained semantic correspondence between video and language in three semantic hierarchies. Experiments illustrate significant improvement of our VISA on compositional generalizability. 
	
	
	\noindent
	 \textbf{Limitations and Futuer work.} We observe some failure cases that VISA cannot discriminate subtle semantics of adverbs, \eg, ``fly close'' to ``fly away''. We expect future research to utilize the new benchmarks to make progress on fine-grained semantics grounding, thus achieving compositional generalization.
	 
	 \noindent
	 \textbf{Acknowledgment.} This work has been supported in part by National Key Research and Development Program of China (2018AAA0101900), Zhejiang NSF (LR21F020004), Key Research and Development Program of Zhejiang Province, China (No.2021C01013), Alibaba-Zhejiang University Joint Research Institute of Frontier Technologies, Chinese Knowledge Center of Engineering Science and Technology (CKCEST), Zhejiang University iFLYTEK Joint Research Center. The author from UCSC is not supported by any of the
	 projects above. We thank all the reviewers for valuable comments. 
	
	{\small
		\bibliographystyle{ieee_fullname}
		\bibliography{egbib}
	}
	
\end{document}